\documentclass{Interspeech2024}
\usepackage[table,xcdraw]{xcolor}

\interspeechcameraready

\title{Optimizing the role of human evaluation in LLM-based spoken document summarization systems}

\name[]{Margaret}{Kroll}
\name[]{Kelsey}{Kraus}

\address{
  Cisco Systems, USA}
\email{makroll@cisco.com, kekraus@cisco.com}


\begin{document}

\maketitle

\begin{abstract}
    




The emergence of powerful LLMs has led to a paradigm shift in abstractive summarization of spoken documents. The properties that make LLMs so valuable for this task -- creativity, ability to produce fluent speech, and ability to abstract information from large corpora -- also present new challenges to evaluating their content. Quick, cost-effective automatic evaluations such as ROUGE and BERTScore offer promise, but do not yet show competitive performance when compared to human evaluations. We draw on methodologies from the social sciences to propose an evaluation paradigm for spoken document summarization explicitly tailored for generative AI content. We provide detailed evaluation criteria and best practices guidelines to ensure robustness in the experimental design, replicability, and trustworthiness of human evaluation studies. We additionally include two case studies that show how these human-in-the-loop evaluation methods have been implemented at a major U.S. technology company.

\end{abstract}

\section{Introduction}

The explosion in popularity of generative AI content has precipitated a race-to-market effect in features offering abstractive summarization of spoken documents. The field has yet to cohere, however, on a central evaluation strategy for AI-generated content. General recommendations on evaluation criteria are available and they evolve as risks of the technology are uncovered \cite{lee2023holistic}. And while fast, inexpensive automatic evaluations are highly-desired, the limitations of existing automated metrics, e.g. ROUGE and BERTScore, have driven the community to pair these metrics with human evaluation tasks, which are still considered the gold standard for LLM-generated content \cite{van-der-lee-etal-2019-best, liu-etal-2023-revisiting, lee2023holistic, zhang2019pegasus, fabbri2021summeval, HELM-2023}.


The use of spoken documents, meeting recordings, and recorded conversations as input media presents an additional challenge to generative AI features. Transcribed speech is subject to a variety of noise, including Automatic Speech Recognition (ASR) mistranscriptions, spelling and grammar inconsistencies, and speaker misattributions. These errors propagate into the final feature output if not corrected prior to passing the input to the generative AI system. Furthermore, standard ASR metrics such as word error rate are not sensitive enough to ensure good quality input, as they undervalue the significance of single word swaps such as speaker name misattributions.

This paper fills a need in the literature regarding best practices for implementing human evaluations of generative AI spoken document summarization. We first propose a suite of evaluation criteria. We then classify each criterion as being appropriate to at least one of four distinct evaluation frameworks: reference-free human evaluation, reference-based human evaluation, reference-free LLM evaluation, or reference-based LLM evaluation. Focusing on the two human-based evaluations, we provide detailed methodological guidelines for implementing evaluations of this class. Two case studies are presented for illustration, in which these methods are used to evaluate two features prior to their release at a U.S technology company. 


\section{Related work}

Several papers have conducted recent meta-analyses of summary evaluation methodologies within NLP.  These papers show that the field suffers from extensive underreporting of experimental designs and evaluation approaches. For example, the majority of papers included in the meta-analyses do not appear to use any form of statistical significance testing to support their claims. Additionally, sample sizes are rarely reported and, when they are, a large majority of experiments are significantly underpowered \cite{van-der-lee-etal-2019-best, card-etal-2020-little, van-der-lee-etal-2020-best}. The main concerns with these practices are twofold. First, failure to report on experimental design renders external evaluation of the researchers' claims impossible, and prevents any efforts to duplicate results. Second, running underpowered studies and/or using inappropriate statistical analyses increases the likelihood of incorrectly rejecting the null hypothesis (Type I error) or failing to reject an incorrect null hypothesis (Type II error), greatly elevating the risk that the researchers are reporting unsubstantiated and/or biased results. The designs proposed here obviate these concerns by outlining detailed methodological best-practices.

There is also a growing body of literature aimed at creating frameworks that use LLMs to assess the quality of generative outputs. ChatEval utilizes several LLMs trained as `agents' to evaluate a task, engaging them in debate that the authors argue leads to better performance than a single LLM evaluation \cite{chan2024chateval}. Wang et al. \cite{wang-etal-2023-chatgpt} compare ChatGPT evaluations of NLG tasks against \textit{n}-gram and embedding-based metrics. The authors explore using ChatGPT in both reference-based (in which the model is presented with a gold summary) and reference-free methods, and conclude that ChatGPT performs competitively with human evaluators in several task types. Another model, called G-Eval, uses GPT-4 and a framework called Chain of Thought to evaluate the quality of LLM-based evaluations of generative output \cite{liu-etal-2023-g, yao2023tree}.  This framework argues that requiring the LLM to output its reasoning along with an evaluation score improves the model's evaluation performance. The conclusions of these studies show that LLM-based evaluations are generally surpassing the quality of \textit{n}-gram and embedding-based metrics as measured by correlations with human evaluation scores. However, the authors claim that LLM evaluation outputs themselves have yet to achieve high correlation with human evaluation scores. Additionally, LLMs show preferential treatment toward the output of other LLMs (and toward the output of other LLMs) over human-created content \cite{liu-etal-2023-g}, and they show high sensitivity to the informational content, wording, and ordering of the prompt content.

\section{Limitations of the ``State of the Art"}

We note two main limitations of existing studies. First, the studies compare ordinal and/or interval scale measurements. While the methodologies are not explained in detail, no paper mentions presenting any practice or burn-in questions to participants, or mentions including a baseline summary for comparison evaluation. However, human judgments, and plausibly model judgments, are context-sensitive and baselines differ across individuals \cite{Hautus_Macmillan_Creelman_2022}.  That is, each rater can and should be expected to have a different baseline for what a `4' rating on a Likert scale is, particularly if no calibration of the scale is performed in the presentation of items. Therefore, we do not necessarily expect high within-participant or between-study agreement.

Furthermore, it is standard in the NLP literature to evaluate the performance of LLM evaluations using correlation scores between LLM-generated scores and human-generated evaluation scores. Generally, three main correlation measurements — Pearson, Spearman, and Kendall’s Tau — are provided. However, not all correlation scores are appropriate in all cases. Pearson correlation, specifically, is a parametric measure that is sensitive to outliers and makes the assumption of normality regarding the distribution of the underlying data. However, no plots or descriptive statistics are provided in the papers reviewed here, and the correlation scores are given without comment on their applicability to the underlying data. 

Regardless, even if the proper correlation measurement is used relative to the data distribution, correlation is a measure of association and not of agreement, neither between-method or between-participant agreement. For example, because correlation is not dependent upon intercept value, two variables with no overlaps in values can achieve a perfect correlation \cite{Watson-2010}. Given this, using correlation as a benchmark to compare the performance of evaluation tasks, especially across datasets and experiments, is fraught if not interpreted carefully \cite{janse-2021-correlation}. Furthermore, even within studies that use an appropriate inner-annotator agreement score for human evaluators, the authors find very low agreement. Low within-participant agreement leads to unreliable assessment of between-methods agreement. The studies additionally contain low numbers of participants (see discussion above), and conduct no hypothesis testing. Therefore, it is challenging to draw valid conclusions about the comparisons made between evaluation methods in the literature.


\section{Evaluation criteria and frameworks}\label{eval-criteria}


A popular emerging use-case of generative AI is summarizing text or spoken documents within collaborative applications. Our work here provide guidance on implementing human evaluation studies for generative AI features that are user-facing in a business environment. We propose four main principles that underly model summary output evaluation \cite{kant}.

\subsection{Evaluation criteria}\label{criteria}

\begin{enumerate}
    \item\label{quantity} \textbf{Quantity}
    \begin{enumerate}
        \item Length: is the output an appropriate length, e.g. minimum and maximum word count?
        \item Recall: is the relevant information from the input returned in the output?
    \end{enumerate}
    \item\label{quality} \textbf{Quality}
    \begin{enumerate}
        \item Precision: is all the information in the output supported by the input (did the model hallucinate)?
        \item Attribution: if content is attributed to a particular speaker, is this attribution correct?
    \end{enumerate}
    \item\label{relevance} \textbf{Relevance}
    \begin{enumerate}
        \item Repetition: does the output repeat information?
        \item Pertinence: does the model output contain information that is irrelevant to the task at hand, e.g. small talk?
    \end{enumerate}
    \item\label{manner} \textbf{Manner}
    \begin{enumerate}
        \item Coherence: does the output contain grammatical errors, dangling anaphors, and/or incoherence language?
        \item Vocabulary: does the vocabulary of the output match the vocabulary of the input?
        \item Tone: does the tone of the output match that of the input?
        \item Organization: is the output organized in a clear and coherent manner?
    \end{enumerate}

\end{enumerate}

\noindent We organize these criteria into four methodological approaches.


\subsection{Reference-free: human-evaluation}

Reference-free evaluations are those in which no gold standard is used. For spoken language document summarization, this method requires evaluation of the summary either with no comparison document, or in comparison to the original audio and/or transcribed text. This design can be time-consuming for participants, and is valuable for obtaining in-depth feedback at the expense of breadth of feedback. For example, comparing a long document against a model-produced summary can be a cognitively demanding and lengthy task. This task is most fruitfully approached in a qualitative, exploratory design.

The benefits of this design is that it does not require golden annotations of documents, and requires fewer participants than a quantitative design. It also provides in-depth feedback on multiple components of the model output. It is additionally compatible with a multi-modal presentation, in which participants watch a video or listen to audio of the meeting. This presentation medium allows the researcher to evaluate properties difficult to capture in transcribed text, such as tone and ASR errors. However, the results from such a study do not generalize to a wider population and do not provide a big-picture evaluation of model performance. The design is also limited in the amount and types of comparisons that can be conducted in one sitting due to carryover and fatigue effects. If a monitored design is used, then each participant is observed by a researcher, which requires that researcher’s time as well as the participants' time. 

This approach allows the researcher to evaluate all the criteria given in Section \ref{criteria}, although it is not a cost-effective approach to use for criteria that are easily automated, such as word-counts.


\subsection{Reference-based: human-evaluation}
 
Reference-based evaluations are those in which a gold summary is used as a ground-truth document for model evaluation. This approach has a rich history in summarization task evaluation, preceding LLM-generated summaries \cite{summac-02, summac-21}. The use of ground-truth documents offers several benefits. For one, the shortened nature of the task means that a much larger sample size can be used for the study. The increased sample size allows the use of multilevel regression models, which include random intercepts and slopes for items and participants. This accounts for differing baselines of individual judgments. In addition, some research has suggested that summary length affects participants’ judgments of the summaries \cite{liu-et-al-23}. This design allows the researcher to include variables like word length as a random regressor, thereby controlling for the potential influence of external factors on participant judgments.

More broadly, this design provides a higher-level, externally valid evaluation of model performance. It requires more participants than the exploratory design, although the survey should take less time per participant. This design requires the creation of gold summaries, which can be time-consuming depending on the nature of the documents. 

We recommend this approach as the primary means to evaluate Quantity (\ref{quantity}) criteria (recall) and Quality (\ref{quality}) criteria (precision). It is also effective to use for Relevance (\ref{relevance}) criteria and Manner (\ref{manner}) criteria.

\subsection{Reference-based: LLM-evaluation}

Human evaluation of LLM-summarization output is resource-intensive. An alternative currently being explored in the literature is using LLMs themselves to evaluate an LLM-produced output. The benefit of such an approach is that it can be scaled in a less resource-intensive way than human evaluations. While we touch on best practices for these evaluations here, we forego an in-depth analysis for future work.

The limitations discussed earlier suggest several design best practices. First, tasks should be created to avoid eliciting ordinal or interval judgments. Yes/No judgments and forced-rankings are two alternative measurements to elicit. Second, if scale judgments must be used, summaries should always be evaluated against a baseline summary, and/or scores must be calibrated in the prompt or instructions. A semi-automated evaluation can also quantify success criteria such as recall/precision of summary content.

Additionally, the LLM should be asked to explain its chain of reasoning for reaching its decision. Preliminary testing also suggests that an LLM performs better when evaluation criteria are defined. For example, aspects such as action items and criteria like fluency should be defined in the prompt. Finally, it is possible to use more expensive but higher-performing models to evaluate the output of a less expensive model. For example, GPT4 can be used to evaluate the recall of the output of GPT3.5.

We recommend this approach for Relevance (\ref{relevance}) and Manner (\ref{manner}) criteria and for Quality (precision). We caution against its use for Quantity metrics (recall) without significant testing.

\subsection{Reference-free: LLM-evaluation}

This design should be approached with caution. It is known that LLMs favor the output of other LLMs, with a bias toward the outputs of their particular model. The risks of evaluating generative AI output using another LLM are therefore high. Even so, there are several criteria that an LLM can plausibly reliably perform. These are criteria that do not rely on ground-truth data, such as word counts and grammatical error detection, or detecting the presence or absence of a particular part of a summary (e.g., there are 4 bullet point notes, but no action items returned). As noted, some of these tasks may be more cost-effective using non-LLM automated methods. It is also possible that a reference-free LLM design can be effectively used to evaluate quality metrics on LLM output. We recommend limiting reference-free LLM approaches to Relevance (\ref{relevance}) and Manner (\ref{manner}) evaluation criteria.

\section{Case studies}

We present two case studies of human evaluation tasks. The first is a reference-free human evaluation, and the second is a reference-based human evaluation. Both were undertaken at a large U.S. tech company as part of a wholistic go-to-market strategy for two voice-based features.

\subsection{Case study: Reference-free human evaluation}

\subsubsection{Methods}

Our first case study evaluates generative AI produced summaries of business meetings. This use case presents a challenge as the content is voice-based and therefore may contain inherited ASR errors, and is often lengthy. The short time-frame of feature development also precluded the creation of gold summaries, which must be created by hand and are therefore time and resource-intensive. The design uses 35 in-house meetings from real business calls and 65 employee recruits as participants.\footnote{Crowdsourcing could not be used due to privacy legal restrictions on the meetings used.} It includes both an overall quality evaluation of the summaries and a model comparison evaluation.


\subsubsection{Materials}

The design was an A/B Latin square presentation with one factor (independent variable), \textit{Summary Type}, containing two levels, \textit{model A} and \textit{model B}. Summaries were randomly selected from a sample of business meetings that took place between July 2022 and January 2024. Summaries were comprised of eight meeting-types – daily standup meetings, feature execution reviews, feature planning meetings, cross-functional team syncs, team-wide all-hands, small working group syncs, weekly reading groups, and cross-team collaborations. A total of 35 summaries were selected for the survey. Each summary was presented in both the A condition (in which the summary was created by generative AI model A), and the B condition (in which the summary was created by generative AI model B). Models A and B differed by prompt details and hyperparameter settings. Due to the Latin Square presentation, participants saw each summary in only one condition. This prevented possible cross-contamination from participants seeing a summary in both conditions.

\subsubsection{Measurements}

Judgments were collected on Quantity, Relevance, and Manner criteria. Fluency and organization were measured using Likert scales from 1-5. Level of detail and unprofessional language were measured using binary Yes/No judgments. A free response option was also given after each survey. The exact metrics as they were defined in the task are given below. 

Participants were trained on metric definitions and were shown an example summary of exemplar quality. As an additional burn-in example, a summary of poor quality was also given. These examples allowed participants to calibrate their scale ranges at the beginning of the task.


\begin{enumerate}
    \item \textit{fluency}: A summary scored as a 5 should be free of any grammatical errors and contain clear and professionally written prose. Summaries scored as less than a 5 may contain things like misspellings, run-on or incomplete sentences, awkward language, or ambiguous phrases.
    \item \textit{organization}: An organized summary does not repeat information or contradict itself, is of a reasonable length for a reader to scan, and it is presented in chronological order with coherent themes.
    \item \textit{level of detail}: Summaries have the right level of detail if they include highlights and action items relevant to the meeting topic, and if they contain enough detail so that each highlight and action item is interpretable. Summaries don’t have the right level of detail if they include idle chit-chat or ice breaker conversations, or if they seem to be missing crucial information.
    \item\textit{unprofessional language}: This language includes, but is not limited to, responses containing instances of hate speech or abusive language, or responses that clearly reflect a personal or evaluative opinion that cannot be derived from factual information.
\end{enumerate}



\subsubsection{Analysis}

We followed best practices for analyzing ordinal (Likert scale) and binomial data. For our ordinal results, we computed median, mean, and standard error of the mean as a measure of central tendency. For the binomial results data, we computed proportions yes/no with 95\% confidence intervals. To determine whether we observed a statistically significant difference between Model A and Model B, ordinal quality metric judgments were analyzed using a cumulative link regression model \cite{ordinal-r} with the factor Model and the levels \textit{model A} and \textit{model B}. To test for a significant difference on the binary judgment data we fit a generalized linear mixed effects model, using glmer with family = binomial(link=logit) \cite{lme4-r}. The factors and levels are identical to the ordinal data.




\subsection{Case-study: Reference-based human evaluation}

\subsubsection{Methods}

Our second case study evaluates a virtual agent chatbot feature. This feature uses Retrieval Augmented Generation (RAG) \cite{rag-21} and was comprised of two components. The first component pulls up-to-date documents from a knowledge base that contains information relevant to a user query. The second component calls an LLM that grounds its response to the user query by summarizing the information retrieved from the knowledge base. RAG systems have become increasingly prominent in the past year due to benefits provided by this first component: the ability to supply up-to-date grounding information to inform and limit the LLM's response. RAG systems are particularly useful in support conversations between a call center agent and a customer. Customer speech is analyzed in real time, and documents relevant to the customer's needs are surfaced in the form of a pre-configured response. 

RAG systems present a challenge for evaluation systems because they require two evaluation frameworks: one quantifying the document retrieval success rate, and the second quantifying the quality of the LLM's response. Using the criteria outlined in Section \ref{eval-criteria}, we gathered human evaluations on a critical subset of our Evaluation Criteria: groundedness in the knowledge base (\ref{quantity}), attention to detail (\ref{quality}), model correctness (\ref{relevance}), and summary appropriateness (\ref{manner}).

\subsubsection{Design Overview}

Participants were shown a screen with a conversation between an agent and a customer that ends in a customer query. Participants were also shown the documents retrieved by the RAG system and two suggested model responses to the customer query: one response from Model A, and one response from Model B. Model conditions were masked from participants, with model responses presented in a random order on screen. 

Participants were given four tasks to perform. The first task framed the LLM's grounding in the knowledge base articles as a classification problem. Participants chose from the options given below for classifying the model's response:

\begin{enumerate}
    \item \textit{True negatives}: The LLM correctly refused to answer because the relevant information was not available in the returned knowledge base articles.
    \item \textit{False positives}: The response contained hallucinated content not in the knowledge base articles.
    \item \textit{False negatives}: The model failed to fetch the relevant information from the knowledge base articles.
    \item \textit{True positives}: The model responsed with correct information from the knowledge base articles.
\end{enumerate}

\noindent This task was evaluated using precision, recall, and F$_1$ scores.

Subsequent tasks became available to participants if  `true positive' was selected for Task 1. These tasks evaluated the overall quality of the LLM's response and provided a direct model comparison score. Task 2 asked the participants whether the summary response sufficiently answered the caller's question using the appropriate level of detail. `No' responses signified that the response contained irrelevant information and/or failed to provide sufficient details to answer the customer’s question. Binary responses were analyzed as proportion yes/no responses with 95\% confidence intervals. Task 3 asked participants to choose which model response -- A or B -- was overall preferred. To test for a significant difference in model preference, we fit a logistic regression  model with Factor Model and levels \textit{model A} and \textit{model B}. Finally, Task 4 provided the participant the opportunity to give free-form annotator feedback on their choices.

\section{Future work}

We aimed here to provide researchers and professionals with detailed guidance on performing human evaluations of generative AI content created from spoken language documents. As generative AI technology continues to involve, so too should our approaches to ensuring quality outputs from these systems. We believe that the next role for human evaluations is to use human evaluation outputs to reinforce automated metrics, either in fine-tuning an evaluation-specific model or as ground-truth for evaluating automated evaluation metrics. While human-in-the-loop evaluation designs may currently reign as the gold standard, the appeal of fast, resource-cheap automated metrics suggests that this area is ripe for innovation. 


\section{Acknowledgements}
We would like to thank Lucien Carroll, Vinay Damodaran, Aishwarya Rao, and Francis Kurupacheril for their contributions to the case studies included in this work.

\bibliographystyle{IEEEtran}
\bibliography{mybib}

\end{document}